%% file: 0_MAIN.tex
\documentclass[10pt,twocolumn,letterpaper]{article}

\usepackage{cvpr}              

\usepackage[table,dvipsnames]{xcolor}
\usepackage{makecell}
\usepackage{tcolorbox}
\usepackage{amsmath}
\usepackage{diagbox}
\usepackage{amssymb}
\usepackage{booktabs}
\usepackage{multirow}
\usepackage{times}
\usepackage{lipsum}
\usepackage{subcaption}
\providecommand{\FullStop}{\text{~\@.\xspace}}
\providecommand{\Comma}{\text{~,\xspace}}

\usepackage{makecell}

\usepackage{xspace}
\usepackage{hhline}   

\usepackage{graphicx}
\graphicspath{{./images/}}

\usepackage[pagebackref,breaklinks,colorlinks,linkcolor=red,urlcolor=red,citecolor=blue,bookmarks=false]{hyperref}

\hypersetup{pdfstartview={FitH}}

\definecolor{amber}{rgb}{1.0, 0.75, 0.0}

\def\paperID{7534} 
\def\confName{CVPR}
\def\confYear{2026}

\begin{document}








\title{SIGMA: A Physics-Based Benchmark for Gas Chimney Understanding\\ in Seismic Images}

\author{Bao Truong$^{1}$, Quang Nguyen$^{1}$, Baoru Huang$^{5,*}$, Jinpei Han$^{2}$, Van Nguyen$^{1}$, \\ Ngan Le$^{3}$, Minh-Tan Pham$^{4}$, Doan Huy Hien$^{1}$, Anh Nguyen$^{5}$\\
{\small $^{1}$FPT Software AI Center}
{\small $^{2}$Imperial College London} 
{\small $^{3}$University of Arkansas} \\
{\small $^{4}$University of South Brittany}
{\small $^{5}$University of Liverpool}
{\small $^{*}$Corresponding author}
\\
{\small \href{https://airvlab.github.io/sigma}{https://airvlab.github.io/sigma}}}


\twocolumn[{%
\renewcommand\twocolumn[1][]{#1}%
\maketitle
\begin{center}
\vspace{-3ex}
  \centering
  \vspace{-1ex}
  \captionsetup{type=figure}
  \Large
\resizebox{\linewidth}{!}{
\setlength{\tabcolsep}{2pt}

\begin{tabular}{c}

\shortstack{\includegraphics[width=\linewidth]{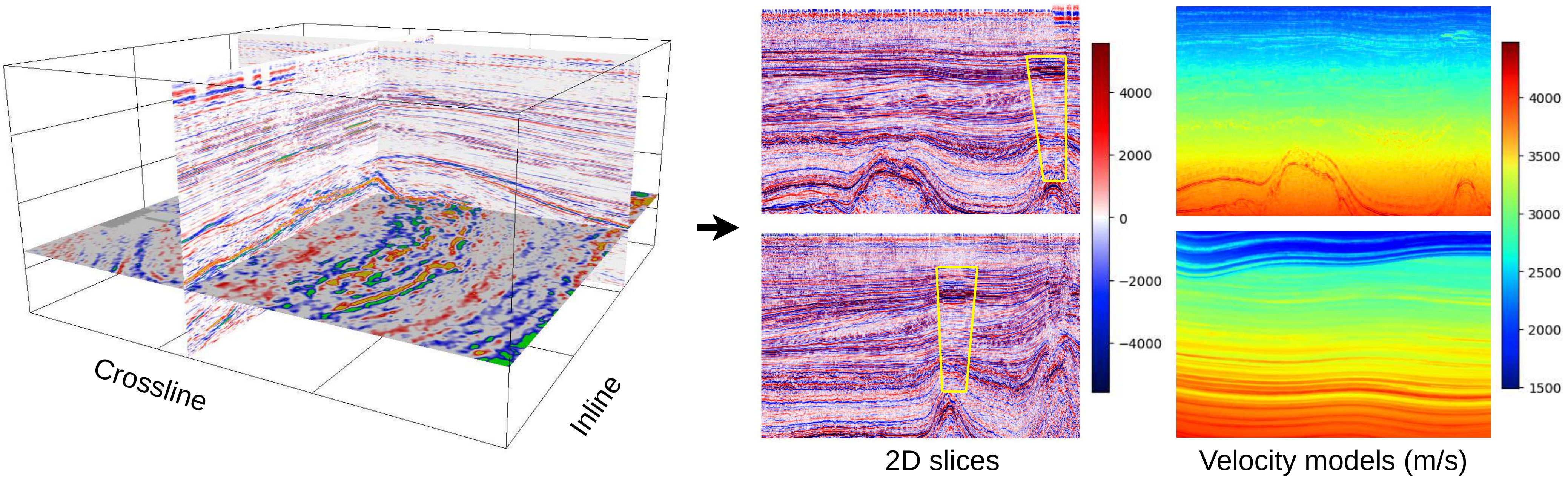}}
\end{tabular}
}
    \vspace{-1ex}
    \captionof{figure}{\textbf{Visualization of a 3D seismic volume.} The 2D seismic slices (expressed in kilometers) are extracted from this volume, with the gas chimney regions highlighted by the \textcolor{amber}{yellow} bounding boxes along with their corresponding velocity models.}
    \label{fig: IntroVis}
\end{center}%
    }]


\input{1_abstract}

\input{2_intro}

\input{3_relatedwork}

\input{4_dataset}

\input{5_proposal_baseline}
\input{7_conclusion}

{\small
\bibliographystyle{ieee_fullname}
\bibliography{egbib}
}

\newpage
\appendix

\end{document}

%% file: 1_abstract.tex
\begin{abstract}
Seismic images reconstruct subsurface reflectivity from field recordings, guiding exploration and reservoir monitoring. Gas chimneys are vertical anomalies caused by subsurface fluid migration. Understanding these phenomena is crucial for assessing hydrocarbon potential and avoiding drilling hazards. However, accurate detection is challenging due to strong seismic attenuation and scattering. Traditional physics-based methods are computationally expensive and sensitive to model errors, while deep learning offers efficient alternatives, yet lacks labeled datasets. In this work, we introduce \textbf{SIGMA}, a new physics-based dataset for gas chimney understanding in seismic images, featuring (i) pixel-level gas-chimney mask for detection and (ii) paired degraded and ground-truth image for enhancement. We employed physics-based methods that cover a wide range of geological settings and data acquisition conditions. Comprehensive experiments demonstrate that SIGMA serves as a challenging benchmark for gas chimney interpretation and benefits general seismic understanding.

\end{abstract}

%% file: 2_intro.tex
\section{Introduction}

Seismic imaging reconstructs views of the Earth’s subsurface by emitting controlled acoustic waves and recording their reflections from geological interfaces~\cite{Yilmaz2001,Claerbout1985}. Unlike traditional visual data, seismic images provide a challenging and largely unexplored type of data where visual structures emerge from physical interactions rather than surface and visual appearance~\cite{robein2010seismic}. The \textit{velocity model} in seismic imaging represents the spatial distribution of seismic wave speeds at each position within the subsurface. An accurate velocity model is crucial for converting seismic data into a meaningful image of the subsurface~\cite{jones2010introduction}. Without a reliable velocity model, the reconstructed seismic images may be distorted or inaccurate, leading to poor interpretations of the geological structures. This model can be estimated from acquired seismic data through Full Waveform Inversion~\cite{schuster2017seismic} and is often used for synthetic seismic data. Fig.~\ref{fig: IntroVis} illustrates the corresponding velocity models alongside their associated seismic slices.

Within seismic volumes, gas chimneys appear as vertically disrupted, low-amplitude, chaotic zones caused by migrating fluids disturbing normal reflection patterns. These features are critical indicators of subsurface fluid flow, offering insight into hydrocarbon charge, seal integrity, and structural anomalies beneath reservoirs~\cite{gay2022future, Ligtenberg2005}. Accurate detection of gas chimneys is essential for exploration success and operational safety, as they often correspond to charged traps or potential drilling hazards~\cite{Reveron2022, zhang2023fault}. In carbon storage settings, pre-existing chimneys may act as vertical leakage conduits, threatening long-term containment~\cite{Chadwick2017}. However, imaging gas chimneys remains a challenging task: their seismic signatures are often subtle, spatially incoherent, and severely degraded by scattering and attenuation~\cite{Ligtenberg2005, zhang2023fault}. Compounding this, chimney labels are rarely available in real data, limiting the ability to validate or train automated detection models~\cite{Reveron2022}.

Conventional seismic workflows rely on physics-based imaging and attribute analysis to detect chimneys~\cite{ismail2020identification}. Advanced techniques like Q-compensated migration~\cite{wang2018adaptive,zhu2014q} can restore attenuation area and improve image fidelity beneath gas-charged zones. However, these pipelines involve costly wave-equation computations and careful tuning of velocity and attenuation models, which may not generalize across different surveys~\cite{Ma2024}. Additionally, real field data provide essentially no ground truth for chimneys. Indeed, there is no record in the public domain of any chimney having been sampled~\cite{Chadwick2017}. Recently, deep learning approaches have emerged for seismic image enhancement and detection~\cite{bargees2023automatic}. 
However, such supervised models require diverse labeled examples; manual annotation of seismic structures in seismic cubes is extremely tedious and subjective~\cite{Wang2025}. This bottleneck motivates our synthetic approach: we introduce \textbf{SIGMA}, a physics-grounded synthetic dataset for chimney understanding. Unlike other seismic synthetic datasets, SIGMA encodes realistic geological variability, structural styles, acquisition geometry, and noise. 

In this paper, we introduce a new large-scale synthetic seismic dataset designed to advance research on gas-chimney–affected seismic image enhancement and detection. The dataset provides (\textit{i}) paired degraded and ground-truth seismic images to facilitate learning-based enhancement and restoration, and (\textit{ii}) pixel-level gas-chimney annotations that enable rigorous detection and segmentation benchmarks. Together, these resources establish a foundation for developing and evaluating both physics-based and deep learning–based methods under controlled, reproducible conditions. In summary, our main contributions are:
\begin{itemize}
    \vspace{-2ex}
    \item We propose SIGMA, a new dataset for seismic image understanding. SIGMA is the first physics-based dataset with ground-truth gas chimney annotations.
    \vspace{-2ex}
    \item We benchmark several baseline methods on gas chimney enhancement and detection tasks, finding that existing approaches struggle on our dataset and generalize poorly to real-world seismic data, revealing open challenges in seismic image understanding.
\end{itemize}

%% file: 3_relatedwork.tex
\section{Related Work}
\textbf{Seismic Image.}     
Seismic images present a rich set of interpretation tasks such as fault detection~\cite{xiong2018seismic}, volcanic detection~\cite{titos2018detection,ibs2008detection}, facies classification~\cite{coleou2003unsupervised,zhao2015comparison, zhao2018seismic}, seismic enhancement~\cite{wang2006inverse,halpert2018deep}, and many other tasks~\cite{wang2019deep, kaur2021seismic,li2019deep,zheng2019applications}. Traditional methods rely on signal processing and seismic attributes (e.g., coherence, semblance, curvature) combined with edge-detection or rule-based algorithms for structure enhancement~\cite{ Hale2013}. Faults and horizons have been mapped using coherence cubes~\cite{bahorich19953}, while salt bodies were typically delineated via thresholding of attribute maps~\cite{amin2019new}. In contrast, recent deep learning methods treat these tasks as semantic segmentation problems, leveraging 2D/3D CNNs, U-Nets, and transformers~\cite{Di2018, Huang2020, Shi2020, lomask2023horizon, araya2017automated, shi2019saltseg, Wu2019}. Horizon tracking~\cite{yang2020seismic}, fault detection~\cite{Wu2019FaultNet}, and facies classification~\cite{Baroni2020} have achieved strong results with limited supervision. Public datasets such as the Penobscot~\cite{Penobscot2016}, F3 Netherlands~\cite{F3}, and TGS Salt Challenge~\cite{TGS2019} provide standardized benchmarks for training and evaluation. Recent works also use synthetic datasets for facies classification task~\cite{xu2023deep}. These advances mark a shift from handcrafted attributes to data-driven representations for seismic structure understanding.

\noindent\textbf{Gas Chimney Effect in Seismic Image.} 
The presence of gas chimneys, or vertical gas migration pathways, poses a significant challenge to conventional seismic imaging and interpretation. These features, characterized by the upward migration of gas from reservoirs, create complex velocity and attenuation anomalies that distort seismic wavefields~\cite{ Ligtenberg2005}. Several imaging techniques~\cite{li2019effective, mao2025high} have been developed to address these challenges. However, their progress critically depends on the availability of suitable seismic datasets. Access to real-world seismic images containing gas chimneys is severely restricted due to the high cost and time required for data acquisition, as well as proprietary constraints imposed by energy companies. Real-world dataset, such as the Netherlands Offshore F3 Block~\cite{F3}, which has 3D seismic survey from the North Sea, is limited by its small scale. Besides real seismic datasets, Arntsen~\textit{et al}.~\cite{arntsen2007seismic} modeled gas saturation to generate synthetic seismic images exhibiting gas chimney effects. However, no existing dataset provides corresponding ground-truth images unaffected by gas chimneys, limiting the development of modern learning-based approaches. 

\begin{figure}[!hbt]
    \centering
    \vspace{-2.5ex}    \includegraphics[width=0.95\linewidth,height=0.7\linewidth]{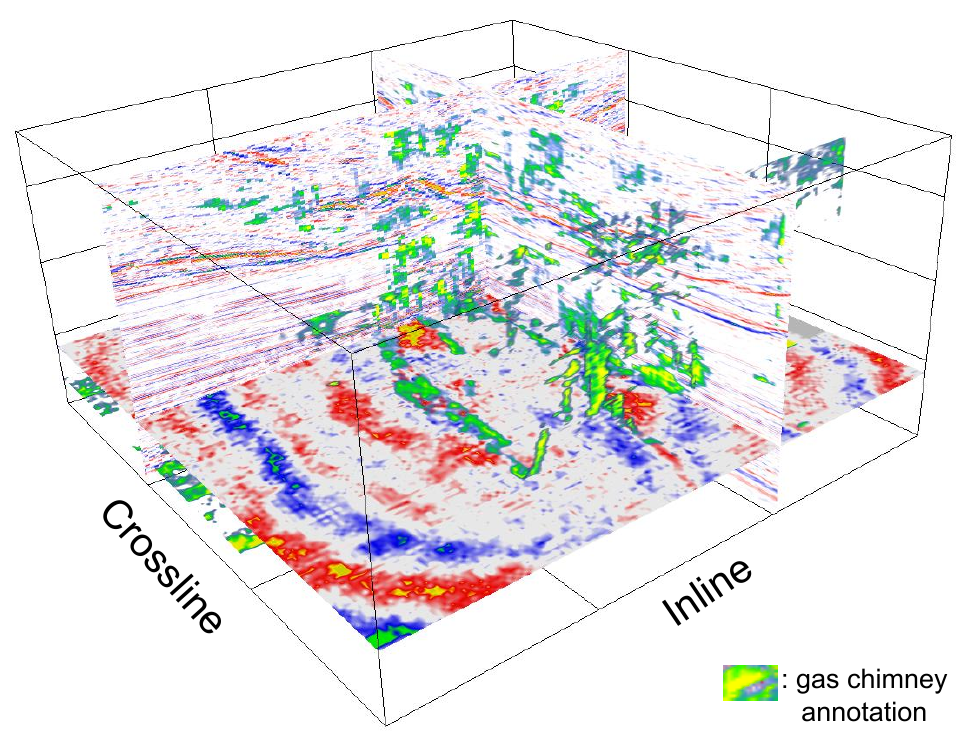}
    \caption{\textbf{An example seismic volume with gas chimney areas.}}
    \label{fig:gas_intro}
\end{figure}

\noindent\textbf{Gas Chimney Detection.} In Fig.~\ref{fig:gas_intro}, we present an illustration of gas chimney detection. Gas leakage can occur widely throughout the subsurface, migrating along fault zones, fractures, or layers. Detecting these features is very challenging because gas chimneys often have irregular shapes, unclear boundaries, and only small differences in contrast compared to the surrounding layers. 
Early gas-chimney studies~\cite{ismail2020identification, el2025gas, imran2021automated, ligtenberg2003chimney, tingdahl2001improving} rely on seismic‐attribute analysis (\textit{e.g.}, amplitudes, phase, frequencies, attenuation similarity-type attributes, etc.) to highlight vertically chaotic, low-coherency zones that indicate fluid migration pathways. Recently, several hybrid approaches~\cite{bargees2023automatic, dixit2020detection, meldahl2001identifying, xu2019semi, xu2017multi} have been proposed that combine physics-based feature extraction with machine learning algorithms to effectively localize chimney channels. These methods typically consist of two stages. Firstly, one or more attributes are extracted from the input data (\textit{e.g.}, recorded signals or migrated seismic images) through physical transformations. Secondly, these features are input into classical machine learning algorithms~\cite{schuster1997bidirectional,gardner1998artificial,hashemi2008gas} to estimate chimney probabilities, which are then used to distinguish chimney from non-chimney regions. Although these methods can capture the multi-faceted seismic expressions of chimneys, they still heavily rely on manual attribute selection, which requires extensive feature engineering and becomes time-consuming when applied to new data. 

\vspace{-2ex}

\noindent\textbf{Seismic Image Enhancement.} In physical terms, the gas chimney effect represents the absorption and scattering of seismic energy, which leads to weak reflections and poor signal continuity during the wave propagation process~\cite{OsorioGranada2023JMSE}. Earlier, several methods have been proposed to estimate and compensate for this energy loss, aiming to reconstruct seismic images under normal conditions, unaffected by gas. The authors in~\cite{ma2022frequency} proposed a stabilized amplitude operator in the frequency domain to estimate the Q factor~\cite{xue2022q}, enabling compensation for attenuation and dispersion effects during the reverse time migration (RTM) process~\cite{Baysal1983}. Other methods~\cite{ning2021q, mao2025high, wang2022explicit, zhou2018efficient, li2019effective} further optimize this process through advanced strategies for Q-factor approximation, aiming to achieve more accurate compensation of seismic attenuation effects. Recent studies such as in~\cite{zhou2022absorption, wang2024enhanced, zhang2022simultaneous, wang2025reconstruction} have attempted to model this quantity using deep neural networks. However, there are still no existing approaches that directly enhance gas chimney–affected seismic images in an image-to-image manner, which remains a significant limitation in this field.

%% file: 4_dataset.tex
\section{Preliminaries}
\vspace{-1ex}
\label{sec_background}

\subsection{Seismic Image Construction Overview}

\vspace{-5ex}
\begin{figure}[hbt]
    \centering
    \includegraphics[width=\linewidth]{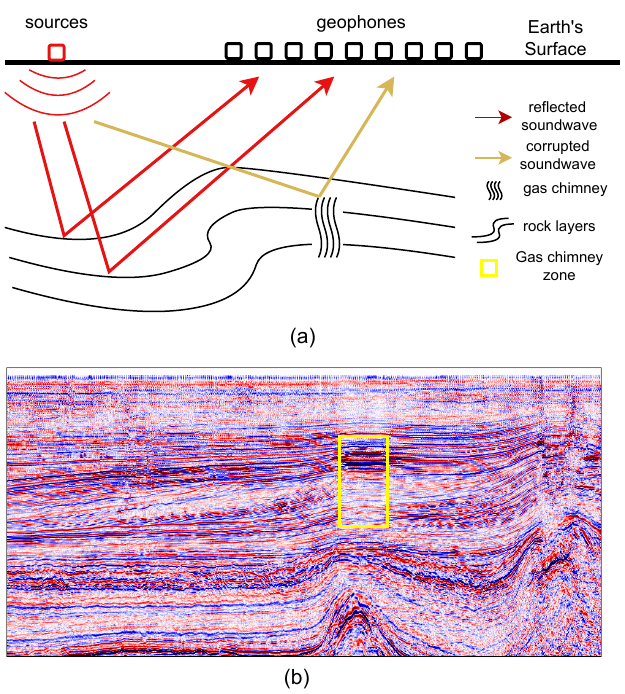} 
     \vspace{-3ex}
    \caption{\textbf{Seismic image construction.} (a) An illustration of seismic image acquisition with gas chimney effect. The sound waves become distorted when they pass through the gas chimney area. (b) The final seismic image after migration, showing that the gas chimney area leads to poor seismic attribute quality.}
    \label{fig:seismic_intro}
\end{figure}
\vspace{-2.5ex}
The overview of seismic image construction is illustrated in Fig.~\ref{fig:seismic_intro}-a. It begins with the generation of acoustic waves from controlled sources at the Earth's surface. These waves propagate through subsurface layers and are partially reflected whenever they encounter contrasts in acoustic impedance, such as between rock interfaces or fluid boundaries. Arrays of geophones or hydrophones positioned at the surface record the returning wavefields, capturing their travel times and amplitudes~\cite{Yilmaz2001}. The acquired traces are then processed through velocity estimation~\cite{schuster2017seismic,li2019deep} to characterize the propagation properties of the medium, followed by migration algorithms such as RTM that back-propagate the recorded signals to reconstruct high-resolution subsurface images \cite{Baysal1983, Claerbout1985}. These seismic images help geoscientists identify faults, traps, and other geological features important for exploration. However, when acoustic waves traverse gas-charged zones, commonly known as gas chimneys, they become significantly distorted and attenuated, leading to blurred or low-coherence regions in the final seismic image (Fig.~\ref{fig:seismic_intro}-b). Such distortion poses a major challenge for accurate subsurface characterization. 

\subsection{Seismic Image Modeling}
\paragraph{Forward Modeling.} Let $\mathbf{x} \in \Omega \subset \mathbb{R}^2$ denote the spatial position and $\tau \in [0,T]$ the time variable. The scalar wave equation~\cite{Claerbout1985,virieux1986p,levander1988fourth} is used to describe the propagation of the scalar wavefield $p(\mathbf{x},\tau)$ in space $\mathbf{x}$ and time $\tau$:
\begin{equation}
    \frac{1}{\mathbf{V}^2(\mathbf{x})} \frac{\partial^2 p(\mathbf{x},\tau)}{\partial \tau^2} - \nabla^2 p(\mathbf{x},\tau) = s(\mathbf{x},\tau)\Comma
    \label{eq:wave_eq}
\end{equation}
where $s(\mathbf{x},\tau)$ is the source term, $\mathbf{V}$ is the velocity model that represents how seismic wave speed changes with depth. Eq.~\ref{eq:wave_eq} is solved forward in time with initial conditions $p(\mathbf{x},0) = 0$ and $\partial_\tau p(\mathbf{x},0) = 0$, together with absorbing boundary conditions to prevent reflections at the model edges. The resulting solution $p_s(\mathbf{x},\tau)$ is called the \textit{forward wavefield} that represents the propagation of energy from the source term through the subsurface.


\paragraph{Reverse Time Migration.} In acoustic approximation, small perturbations in the subsurface can be modeled using the Born approximation~\cite{bleistein2001mathematics}.  
Let the squared slowness be $m(\mathbf{x}) = m_0(\mathbf{x}) + \delta m(\mathbf{x})$, where $m_0(\mathbf{x}) = 1/\mathbf{V}_0^2(\mathbf{x})$ is a smooth background model and $\delta m(\mathbf{x})$ denotes small reflectivity perturbations. Linearizing the wave equation yields the scattered wavefield:
\begin{equation}
\label{eq:born}
\frac{1}{\mathbf{V}_0^2(\mathbf{x})} \frac{\partial^2 \delta p(\mathbf{x},\tau)}{\partial \tau^2} - \nabla^2 \delta p(\mathbf{x},\tau)
= -\,\delta m(\mathbf{x})\,\frac{\partial^2 p_s(\mathbf{x},\tau)}{\partial \tau^2},
\end{equation}
where $p_s(\mathbf{x},\tau)$ is the incident (source) wavefield in the background model.  
This shows that reflections arise from the interaction between the model perturbation $\delta m$ and the curvature of the propagating wavefield.

Reverse time migration (RTM) reconstructs $\delta m(\mathbf{x})$ by back-propagating the recorded data through the same background model and correlating it with the forward wavefield~\cite{whitmore1983iterative,li2021angle}.  
The receiver (backward) wavefield $p_r(\mathbf{x},\tau)$ is obtained by solving:
\begin{equation}
\label{eq:backward}
\frac{1}{\mathbf{V}_0^2(\mathbf{x})} \frac{\partial^2 p_r(\mathbf{x},\tau)}{\partial \tau^2} - \nabla^2 p_r(\mathbf{x},\tau)=\sum_{r} d(\mathbf{x}_r, T-\tau)\,\delta(\mathbf{x} - x_r),
\end{equation}
where $d(\mathbf{x}_r,\tau)$ are the recorded traces and $T$ is the recording time, $\delta(\cdot)$ is the Dirac delta function and $T$ is the total recording time.
The reflectivity image is then formed using the zero-lag cross-correlation imaging condition:
\begin{equation}
\label{eq:rtm}
I(\mathbf{x}) = \int_0^T p_s(\mathbf{x},\tau)\,p_r(\mathbf{x},\tau)\,\mathrm{d}\tau \FullStop
\end{equation}

The reverse time migration thus applies the adjoint of the Born operator, mapping scattered data back into the model space~\cite{symes2007reverse,zhang2003true}. Although not a true inverse, it effectively highlights reflectors and structural boundaries, with image quality largely depending on the accuracy of the background velocity model $\mathbf{V}_0(\mathbf{x})$.

\begin{figure*}[hbt]
    \centering
    \includegraphics[width=\linewidth]{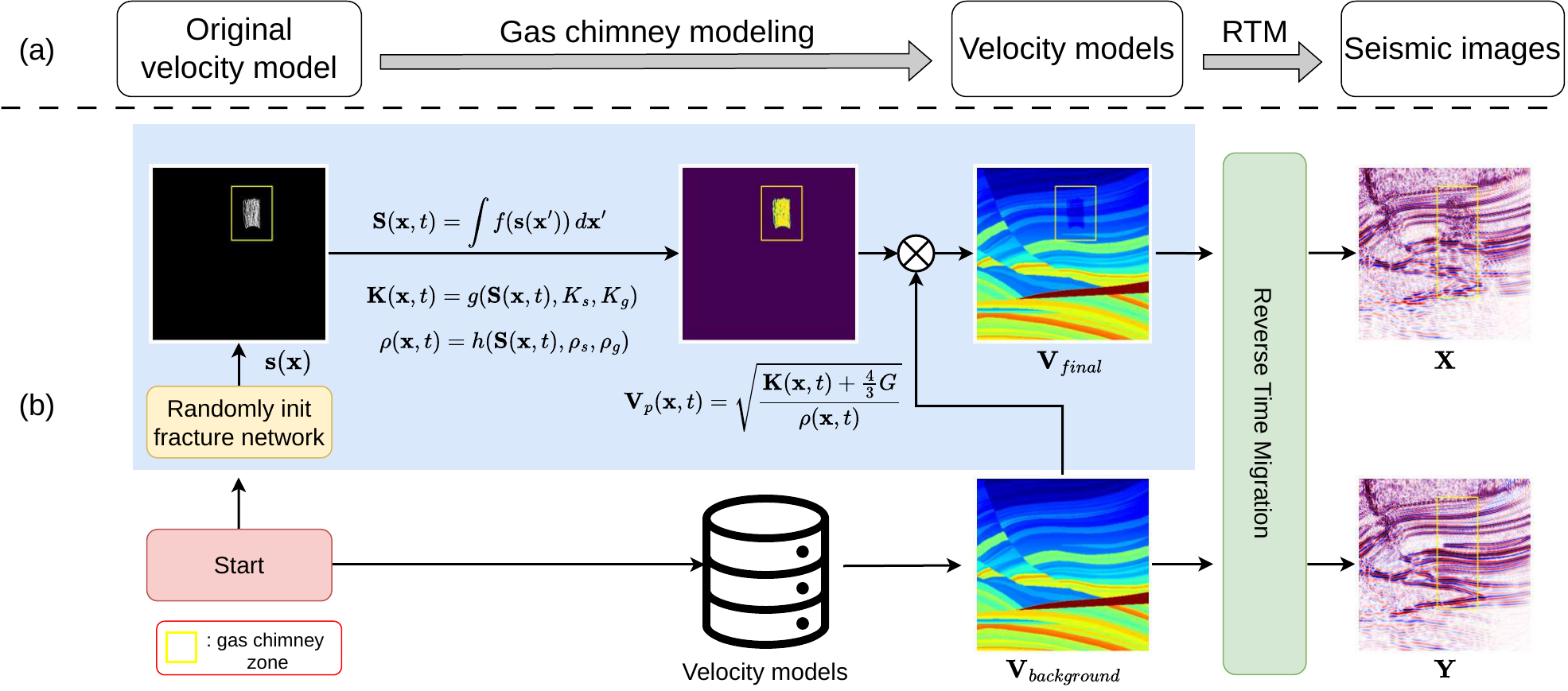}
    \vspace{-3ex}
    \caption{\textbf{Dataset creation pipeline.} (a) Overview of seismic image construction framework with gas chimney. (b) Step-by-step dataset generation process: starting from an original velocity model, a random fracture network is generated, followed by gas saturation simulation through a physically grounded modeling stage. The gas-affected velocity model is then constructed, and final seismic images are synthesized using reverse time migration.}
    \vspace{-2ex}
    \label{fig:data_pipeline} 
\end{figure*}

\section{The SIGMA Dataset}



In the real-world, obtaining labeled seismic images is extremely challenging because it requires direct information from the subsurface, which can only be obtained through costly and time-consuming drilling operations. 
To fill this gap, we introduce a physics-based synthetic pipeline designed to generate realistic seismic images with the ground truth/label pairs tailored for deep learning model training. Our dataset comprises original velocity models, clean seismic images, gas saturation maps, and the corresponding modified velocity models and seismic images with gas chimney effects. 
\subsection{Seismic Images with Gas Chimney Generation}

The proposed dataset generation pipeline is illustrated in Fig.~\ref{fig:data_pipeline}. Initially, a clean velocity model is modified following the acoustic model described in~\cite{arntsen2007seismic} to incorporate the gas chimney effect. Subsequently, RTM is applied to transform the velocity model into the corresponding seismic image. The acoustic model shows gas diffusing from the fracture network into the surrounding rock, causing uneven gas saturation that distorts seismic waves. To generate the fracture network, we first pick random starting points. From each point, a fracture is grown using a random-walk process, where its size is limited by a maximum length and its orientation is restricted to a range of angles relative to the vertical. The fracture network is presented as a function $s(\mathbf{x})$, where $s(\mathbf{x})=1$ if position $\mathbf{x}$ lies within a fracture or $s(\mathbf{x})=0$ otherwise. The gas volume fraction $\mathbf{S}(\mathbf{x},t)$ at position $\mathbf{x}$ and time $t$ is given by:
\begin{equation}
    \mathbf{S}(\mathbf{x},t) \;=\; \int d\mathbf{x}' \, s(\mathbf{x}') \, 
\frac{H_0}{8 \pi D R} 
\left[1 - \mathrm{erf}\!\left(\frac{R}{4Dt}\right)\right]\Comma
\end{equation}
where $R^2 = (x - x')^2 + (z - z')^2$, $H_0$ is the injection rate of gas into the shale (per unit time per fracture point), $D$ is the diffusion constant, $\mathrm{erf}(\cdot)$ is the error function:
\begin{equation}
\mathrm{erf}(\mathbf{x}) = \frac{2}{\sqrt{\pi}} \int_{0}^{\mathbf{x}} e^{-u^2} \, du\FullStop
\end{equation}
The effective bulk modulus of the shale formations is changed due to the presence of gas:
\begin{equation}
    \frac{1}{\mathbf{K}(\mathbf{x},t)} \;=\; 
\frac{1 - \mathbf{S}(\mathbf{x},t)}{K_s}
+ \frac{\mathbf{S}(\mathbf{x},t)}{K_g}\Comma
\end{equation}
where $K_g$ is the gas bulk modulus and $K_s$ is the bulk modulus of the shale formations. The density is also changed by gas saturation and given by:
\begin{equation}
    \rho(\mathbf{x},t) \;=\; (1 - \mathbf{S}(\mathbf{x},t)) \, \rho_s + \mathbf{S}(\mathbf{x},t) \, \rho_g \Comma
\end{equation}
where $\rho_s$ is shale density, $\rho_g$ is gas density. The bulk modulus and density of gas based on pore pressure and temperature can be calculated following the method in~\cite{batzle1992seismic}. Finally, we can obtain the compressional-wave velocity:
\begin{equation}
\begin{aligned}
\mathbf{V}_p(\mathbf{x},t) &= \sqrt{\frac{\mathbf{K}(\mathbf{x},t) + \tfrac{4}{3}G}{\rho(\mathbf{x},t)}}\Comma \\
\end{aligned}
\end{equation}
where $G$ is the shear modulus, and is not dependent upon the fluid properties. The final velocity model with gas chimney effect is obtained by: 
\begin{equation}
    \mathbf{V}_{final} = \mathbf{V}_p * \mathbf{V}_{background} \FullStop
\end{equation}

Consequently, using Equations~\ref{eq:born},~\ref{eq:backward}, and~\ref{eq:rtm}, we construct the clean seismic image $\mathbf{Y}$ from the background velocity model $\mathbf{V}_{background}$, and the gas-affected seismic image $\mathbf{X}$ from the gas-affected velocity model $\mathbf{V}_{final}$. The configurations used for forward seismic modeling are detailed in Table~\ref{tab:gcd_physical_config}. Each sample in our dataset is defined as:
\begin{equation}
\mathbf{S} = \{ \mathbf{X}, \mathbf{Y}, \mathbf{V}_{background}, \mathbf{V}_{final}, \mathbf{V}_p\} \FullStop
\end{equation}


\begin{figure*}[hbt]
    \centering
    \includegraphics[width=\linewidth]{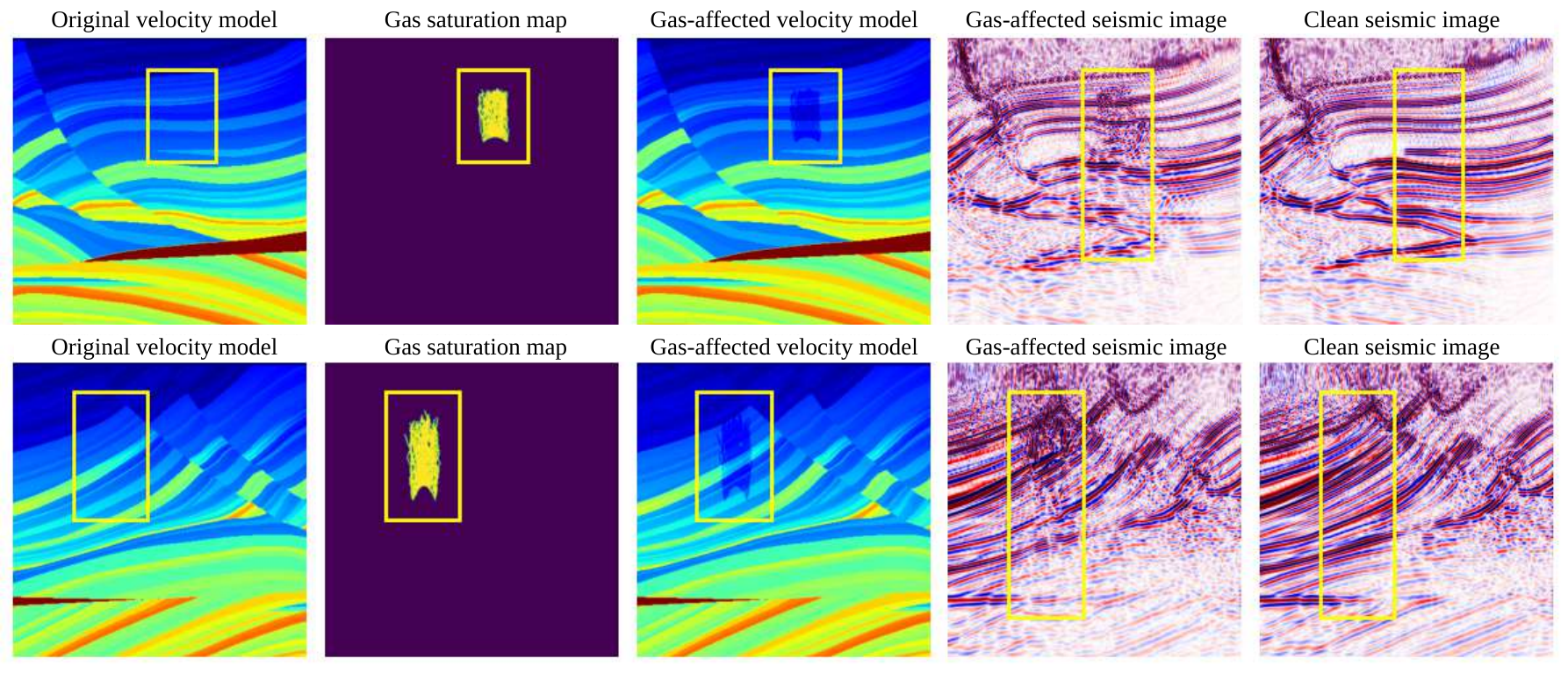}
    \caption{\textbf{Data samples.} We present samples from our SIGMA, each sample consists of original velocity model, gas saturation map, gas-affected velocity model, gas-affected seismic image and clean seismic image. The yellow bounding boxes highlight the differences between normal and gas-affected regions.}
    \label{fig:data_samples}
\end{figure*}

\begin{table}[hbt]
\centering
\vspace{4pt}
\resizebox{\columnwidth}{!}{
\begin{tabular}{l l l l}
\toprule
\textbf{Category} & \textbf{Parameter} & \textbf{Value} & \textbf{Metric} \\
\midrule
\multirow{11}{*}{\textbf{\makecell[l]{Gas \\ chimney\\ modeling}}} 
& Gas bulk modulus ($K_g$) & $8 \times 10 ^{8}$ & Pa \\
& Shale bulk modulus ($K_s$) & $0.045 \times 10^{9}$ & Pa \\
& Shear modulus ($G$) & $3211$ & Pa \\
& Initial gas density ($\rho_g$) & $1900$ & $\text{kg/m}^3$ \\
& Shale density ($\rho_s$) & $1900$ &$\text{kg/m}^3$ \\

& Diffusion time ($t$) & $750$ & $year$ \\
& Diffusion coefficient ($D$) & $10^{-12}$ & $\text{m}^2\text{s}^{-1}$ \\
& Gas supply rate ($H_0$) & $(8\pi D)^{-1} \times 10^{-3}$ & -- \\
& Pore pressure ($p$) & $\rho_w g z$ & Pa \\
& Water density ($\rho_w$) & $1000$ & $\text{kg/m}^3$ \\
& Gravitational acceleration ($g$)& $9.81$ & $\text{m/s}^2$ \\
& Depth coordinate ($z$) & Variable & m \\
\midrule
\multirow{8}{*}{\textbf{\makecell[l]{Forward \\ modeling}}} 
& Grid spacing & $4$ & m \\
& Source frequency & $15$ & Hz \\
& Source spacing & $80$ & m \\
& Source numbers & $25$ & -- \\
& Receiver spacing & $20$ & m \\
& Receiver numbers & $100$ & -- \\
& Time spacing & $0.004$ & s \\
& Time steps & $511$ & -- \\
\midrule
\multirow{5}{*}{\textbf{\makecell[l]{SIGMA \\ dataset \\properties}}} 
& Grid size & $512 \times 512$ & cell \\
& Survey area & $2.048 \times 2.048$ & km$^2$ \\
& Source line length & $2.0$ & km \\
& Receiver line length & $2.04$ & km \\
& Recorded time & $2.044$ & s \\
\bottomrule
\end{tabular}}
\caption{\textbf{Simulation and modeling parameters.} Physical constants, hyperparameters, and numerical configurations used in our SIGMA dataset generation framework.}
\vspace{-3.5ex}
\label{tab:gcd_physical_config}
\end{table}

\subsection{Data Labeling}\label{subsec: data-labeling}

\begin{figure}[!hbt]
    \centering
    \includegraphics[width=\linewidth]{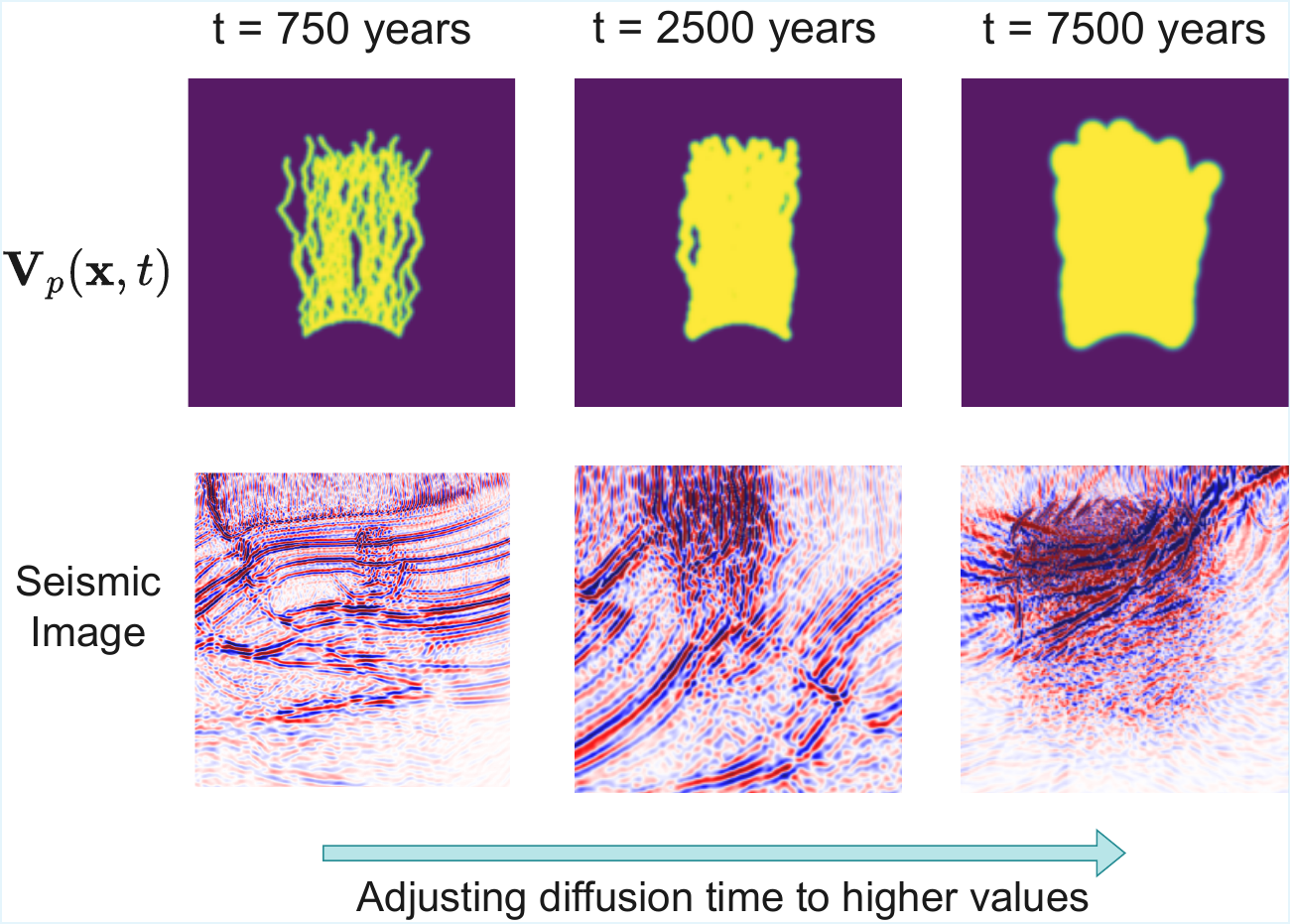} 
    \caption{\textbf{Effect of different physical conditions.} Visualization of variations in the gas chimney effect $\mathbf{V}_p(\mathbf{x},t)$ and their impact on the results under different physical constant settings.}
    \label{fig:param_effect}
    \vspace{-3ex}
\end{figure}

We employed Deepwave~\cite{richardson_alan_2025}, a framework that simulates seismic wave propagation and implements RTM to generate seismic images from velocity models. Both the original velocity model and a modified velocity model incorporating the gas chimney effect were processed with Deepwave to produce corresponding seismic images. In Fig.~\ref{fig:data_samples}, we present the original and modified velocity models alongside their corresponding seismic images. In Fig.~\ref{fig:param_effect}, we visualize how different parameter choices affect the output. The seismic texture changes more noticeably when increasing physical factors such as time diffusion or gas supply rate. 



\subsection{Data Statistics}
We collect 20 real-world velocity models with different resolutions and exhibiting diverse geological structures and spatial patterns~\cite{segseam, geoazur_wind_data}. For each velocity model, 20 fracture networks with randomly assigned locations and sizes. To generate a synthetic sample of size \(512 \times 512\) pixels, we utilized one NVIDIA RTX 8000 GPU with nearly 30~GB of VRAM, requiring about 30-45~minutes for the full simulation and optimization process for one sample. This also highlights the extremely high computational cost in this domain. Each sample in our dataset covers an area of more than 4.2 km$^2$ in real-world conditions. Collectively, our dataset has 400 seismic image pairs, spanning over 1,600 km$^2$, a vast area that is large enough to support gas chimney surveys on a global scale. 


\begin{figure}[h]
    \centering
    \includegraphics[width=\linewidth]{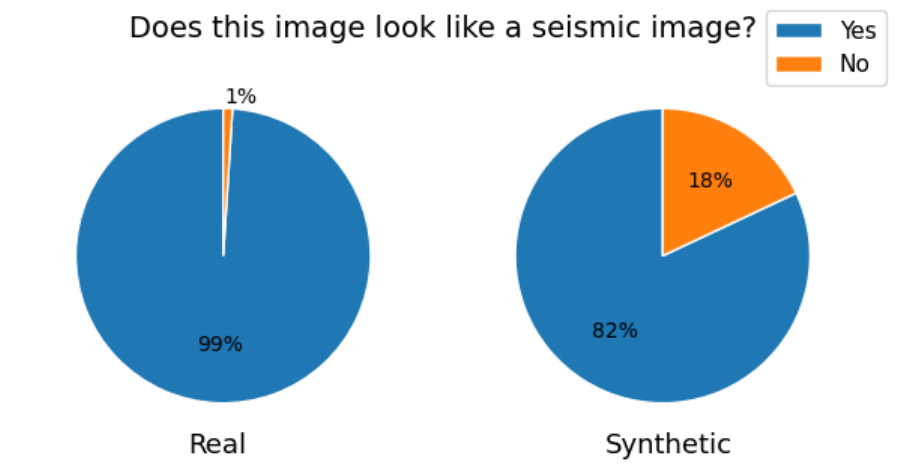}
    \caption{\textbf{User study for assessing realism of synthetic seismic images.} We conducted a perceptual test where each participant was shown ten unlabeled seismic images (five real and five synthetic) in randomized order and asked to judge whether each image \emph{looks like a seismic image} or \emph{not}. The chart summarizes the overall response distribution across ground-truth categories, reflecting how realistic the synthetic images appear compared to real ones.}
    \label{fig:user_study}
    \vspace{-3ex}
\end{figure}
\subsection{Dataset Validation}
To evaluate the perceptual realism of the generated seismic images, we conducted a user study involving participants who are AI researchers or geologists working in the seismic imaging domain. Each participant was presented with ten shuffled and unlabeled seismic samples (five real and five synthetic) and asked to determine whether each image appeared to be a real seismic image. In total, we collected 80 participants, recorded 800 responses (400 for synthetic and 400 for real). To ensure that the perceptual realism estimate is statistically reliable, we accounted for the fact that each participant rated multiple images by applying a cluster-based variance adjustment using an assumed upper-bound intra-class correlation of 0.10~\cite{koo2016guideline}. This increases the design effect to 1.4 and yields an effective sample size of approximately 286 instead of 400. Using this effective sample size, we treat the proportion of ``Yes'' responses (82\% for synthetic images, as shown in Fig.~\ref{fig:user_study}) as a binomial proportion. The corresponding 95\% Wilson confidence interval is $[0.78,\,0.86]$ (indicating a small margin of error $\pm 4.0\%$). To understand the 18 \% negatives on synthetic images, we aggregated responses at the sample level. We found that almost negative responses were concentrated in a specific sample. Qualitative inspection of this sample reveals a typical failure case (where seismic textures are blurred, collapsed, as a sample shown in Fig.~\ref{fig:param_effect}).  This finding suggests that the 18\% ``No'' reflects a few degraded samples rather than widespread low realism. These analysis results demonstrate that the synthetic images capture key structural and textural characteristics of real seismic signals, thereby validating the effectiveness of the proposed framework.

%% file: 5_proposal_baseline.tex
\section{Tasks and Benchmarks}
In this section, we benchmark the SIGMA dataset on primary tasks that are critical in real-world applications: (\textit{i}) localization of the gas chimney regions, and (\textit{ii}) enhancement of seismic features in the gas chimney area. We divide our dataset into a training and a testing set with no overlap in velocity models. The testing set was generated from 5 velocity models, corresponding to 100 seismic image pairs, while the training set consisted of the remaining 15 velocity models, corresponding to 300 seismic image pairs.

\newcolumntype{L}[1]{>{\raggedright\arraybackslash}p{#1}}
\newcolumntype{C}[1]{>{\centering\arraybackslash}p{#1}}
\newcommand{\up}{\ensuremath{\uparrow}}
\newcommand{\down}{\ensuremath{\downarrow}}
\newcommand{\na}{\textemdash}        
\newcommand{\best}[1]{\textbf{#1}}   

\begin{table}[h]
  \centering
  \small
  \setlength{\tabcolsep}{4.5pt}
  \renewcommand{\arraystretch}{1.1}
  \begin{tabular}{
    L{2.2cm}   
    C{0.9cm}   
    C{1.5cm}  
    C{1.5cm}  
  }
  \toprule
  \textbf{Model} & \textbf{Year} &  \textbf{IoU\,\up} & \textbf{Dice\,\up} \\
  \midrule
  FaultSeg~\cite{wu2019faultseg3d} & 2019  & 0.76 & 0.86 \\
  DualUnet~\cite{wang2023transformer} & 2023  & 0.84 & 0.91 \\
  FaultFormer~\cite{wang2025attentionfaultformer} & 2025 & 0.80 & 0.87 \\
  FaultViT~\cite{li2025faultvitnet} & 2025  & 0.75 & 0.86 \\
  \bottomrule
  \end{tabular}
    
    \caption{\textbf{Gas chimney detection results.} 
    }
    \label{tab:detection_benchmark}
    \vspace{-4ex}
\end{table}

\vspace{0.5ex}

\vspace{0.5ex}
\subsection{Gas Chimney Detection}
\textbf{Problem description.} Gas chimney detection aims to localize the gas reservoirs in faulty regions within the seismic images. This is a crucial task in seismic image reconstruction, as it serves as a preliminary step for subsequent interpretation, processing, and enhancement stages. In this work, we formulated this task as a segmentation problem, where the model predicts the pixel-level masks corresponding to chimney structures within the seismic image.   

\begin{figure*}
    \centering
    \includegraphics[width=\linewidth]{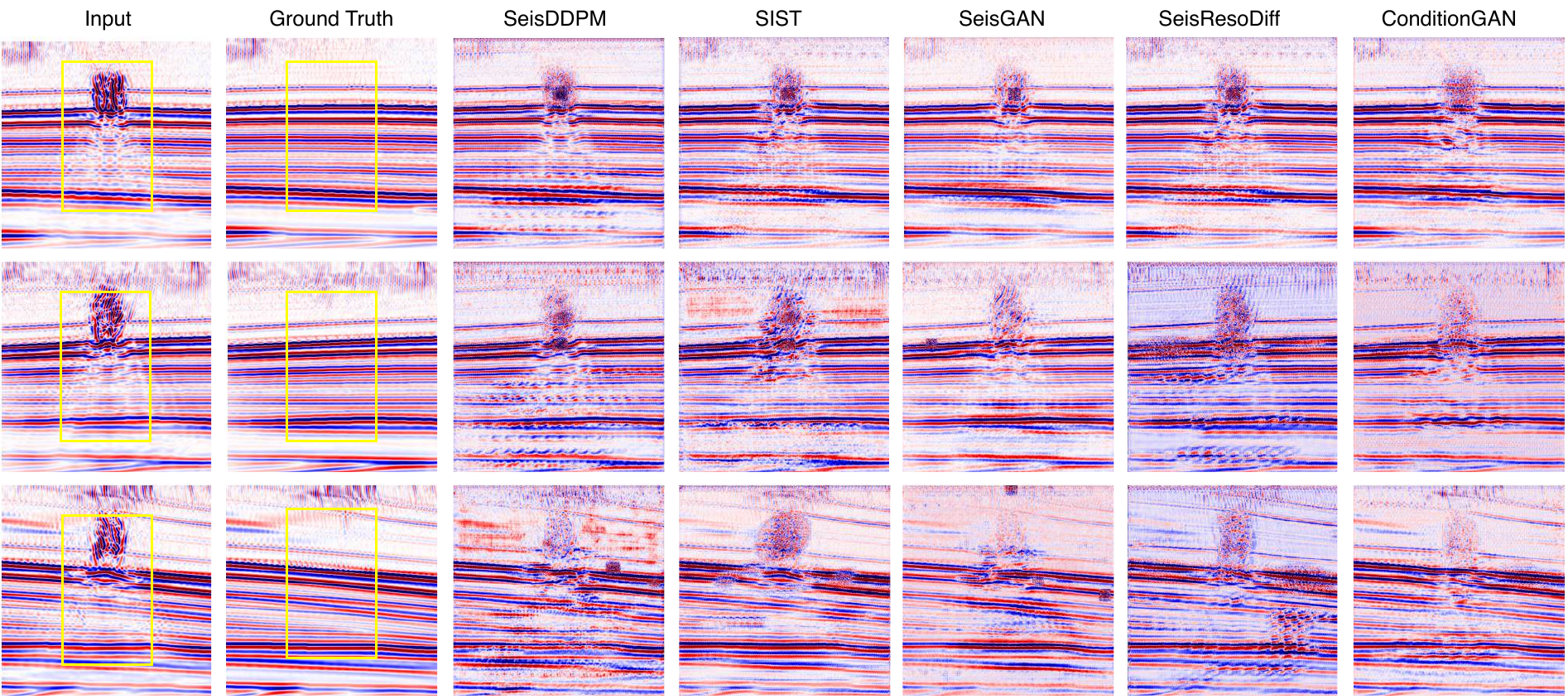}
    \caption{\textbf{Qualitative on enhancement task.} We compare gas chimney enhancement outputs from different methods. The yellow bounding boxes highlight degraded gas chimney regions where models are expected to enhance and reconstruct seismic features.}
    \label{fig:exps_enhancement}
\end{figure*}

\vspace{0.5ex}
\noindent\textbf{Baselines and Training.} We evaluate four recent methods: a CNN-based model (FaultSEG~\cite{wu2019faultseg3d}), a dual-path U-net variant (DualUnet~\cite{wang2023transformer}), and two recent Transformer-style models (FaultFormer~\cite{wang2025attentionfaultformer} and FaultViT~\cite{li2025faultvitnet}). All models are trained using three NVIDIA RTX 8000 GPUs. We adopt the Dice loss function~\cite{milletari2016v} and the Adam optimizer with a momentum factor of $0.98$ and an initial learning rate of $0.001$. A cosine annealing scheduler is employed to gradually reduce the learning rate during training, improving model convergence and stability.

\noindent\textbf{Evaluation and Results.} To evaluate detection performance, we employ Intersection over Union (IoU) and Dice Coefficient, as summarized in Table~\ref{tab:detection_benchmark}. While existing methods demonstrate promising results, the task of accurately detecting gas-saturated regions remains critical and complex, as it is a crucial process for subsequent enhancement methods. The importance of achieving high accuracy in gas chimney detection cannot be overstated. Thus, while current methods have shown promising performance, the complexity and high-risk nature of this task demand further improvements in feature representation and modeling strategies to overcome the inherent challenges of seismic images and ensure that gas chimney detection reaches the level of precision required for practical, real-world applications.


\vspace{1ex}
\subsection{Gas Chimney Enhancement}
\vspace{1ex}
\textbf{Problem description.} Given a 2D seismic image with a corresponding gas chimney mask, our goal is to interpret the collapsed gas reservoir area and restore the seismic texture and structural continuity within these regions degraded by the gas-related effects.

\begin{table}[h]
\vspace{-1ex}
  \centering
  \small
  \setlength{\tabcolsep}{4.5pt}
  \renewcommand{\arraystretch}{1.1}
  \begin{tabular}{
    L{2.4cm}  
    C{0.8cm}  
    C{0.9cm}  
    C{0.9cm}  
    C{0.8cm}
    C{0.7cm}
  }
  \toprule
  \textbf{Model} & \textbf{Year} & \textbf{SSIM\,\up} & \textbf{PSNR\,\up} & \textbf{Corr\,\up} &
  \textbf{SNR\,\up}
  \\
  \midrule
ConditionGAN~\cite{Zhang2022CGAN} & 2022 & 0.52 & 20.02 & 0.81 & 3.33 \\ 
SeisDDPM~\cite{Durall2023DiffusionSeismic} & 2023 & 0.41 & 16.14 & 0.61  & 0.68\\
  SeisGAN~\cite{lin2023seisgan} & 2023 & 0.30 & 15.66 & 0.46 & 2.08 \\
  SeisResoDiff~\cite{Zhang2024SeisResoDiff} & 2024 & 0.30 & 16.05 & 0.70  & 0.59 \\

  SIST~\cite{Dong2024SIST} & 2024 & 0.65 & 19.00 & 0.77 & 3.51  \\
  
  \bottomrule
  \end{tabular}
   \caption{\textbf{Gas chimney enhancement results.} 
   }
   \label{tab:enhancement_benchmark}
   \vspace{-4ex}
\end{table}

\vspace{1ex}
\noindent\textbf{Baselines and Training.} We benchmark five recently network architectures: two diffusion-based models (SeisResoDiff~\cite{Zhang2024SeisResoDiff}, SeisDDPM~\cite{Durall2023DiffusionSeismic}), two adversarial models (SeisGAN~\cite{lin2023seisgan}, ConditionGAN~\cite{Zhang2022CGAN}), and a learning-based enhancement model SIST~\cite{Dong2024SIST}. All models are trained under a unified protocol with standard normalization and augmentation to ensure fair comparison. For GAN-based models, we adopt the  Wasserstein Generative Adversarial Network with Gradient Penalty (WGAN-GP) objective for stable adversarial optimization, with the default setup as described in~\cite{lin2023seisgan, Zhang2022CGAN}. Diffusion models are trained using the standard DDPM~\cite{ho2020denoising} framework with noise reconstruction Mean Squared Error (MSE) loss. The learning-based model follows the same training strategy as used in the gas chimney detection task with MSE loss function.

\noindent\textbf{Evaluation and Results.} 
We evaluate four metrics for this enhancement task, consisting of Structural Similarity Index (SSIM), Peak Signal-to-Noise Ratio (PSNR), Correlation Coefficient (Corr), and Signal-to-Noise Ratio (SNR). As shown in Table~\ref{tab:enhancement_benchmark}, the performances of five methods are quite close and all achieve a relatively low SSIM and SNR score, which indicates that gas chimney enhancement is still a very challenging task. By comparing with ground truth in the second column of Fig.~\ref{fig:exps_enhancement}, we observe that all methods still struggle to reconstruct seismic features within and beneath the gas chimney regions, features that are very crucial for supporting geological interpretation and decision-making. These findings highlight a new challenge for the community and call for further research in this direction.



%% file: 7_conclusion.tex
\section{Discussion}

\textbf{Broader Impact.} 
Apart from gas chimney understanding tasks, we highlight several promising research directions that can benefit from our dataset:
\begin{itemize}
    \item \textbf{Seismic Image Interpolation:} While existing methods~\cite{yuan2022self, huang2022self, goyes2025cddip, kaur2021seismic, wang2019deep} primarily focus on interpolating seismic traces or features from conventional datasets, our dataset introduces new and diverse features derived from gas-chimney modeling, which is a phenomenon not typically captured in traditional datasets. These unique characteristics make it valuable for advancing general-purpose seismic image interpolation. We believe that the release of our dataset will further encourage research in this direction.
    \item \textbf{Seismic Image Foundation Model:} Beyond the sole prior work~\cite{sheng2024seismic}, developing a general large-scale seismic foundation model requires a comprehensive understanding of diverse seismic phenomena and also diverse seismic datasets. Our dataset contributes to this goal by simulating gas-chimney occurrence, an important and challenging aspect of seismic imagery, thereby serving as a valuable resource for learning rich seismic representations and improving generalization in seismic foundation models.
\end{itemize}

\noindent\textbf{Limitation.}
Although our proposed dataset helps bridge the gap caused by the lack of annotated gas chimney seismic image datasets for data-driven methods, it still has certain limitations. In the seismic imaging domain, there remains a crucial need for well-defined 3D seismic volumes containing gas chimney structures. We consider this an important direction for future work. In the current work, data generation still suffers from high computational costs, as detailed in Section~\ref{subsec: data-labeling}, which also limits the available resources and their broader applicability. More efficient algorithms are expected to improve in this direction.

\noindent\textbf{Conclusion.} We have introduced SIGMA, the first physics-based dataset with gas chimney annotations. The dataset is generated through physically grounded simulations that capture diverse geological settings and chimney morphologies, reflecting real-world conditions, and enabling training and evaluation of data-driven seismic analysis models. We believe that the release of this dataset will promote further research in automated seismic interpretation and subsurface anomaly detection related to gas chimney occurrences.